\documentclass{article}

\usepackage{PRIMEarxiv}
\usepackage{float} % 导入宏包
\usepackage[utf8]{inputenc} % allow utf-8 input
\usepackage[T1]{fontenc}    % use 8-bit T1 fonts
\usepackage{hyperref}       % hyperlinks
\usepackage{url}            % simple URL typesetting
\usepackage{booktabs}       % professional-quality tables
\usepackage{amsfonts}       % blackboard math symbols
\usepackage{nicefrac}       % compact symbols for 1/2, etc.
\usepackage{microtype}      % microtypography
\usepackage{lipsum}
\usepackage{fancyhdr}       % header
\usepackage{graphicx}       % graphics
\graphicspath{{media/}}     % organize your images and other figures under media/ folder

\title{Det-SAM2: Self-Prompting Segmentation Framework Based on Segment Anything Model 2}

\author{
 Zhiting Wang \\
  \texttt{wangzt@motern.com} \\
   \And
 Qiangong Zhou \\
  \texttt{zhouqg@motern.com} \\
  \\
  Shenzhen Motern Technology Co., Ltd.
  \And
 Zongyang Liu \\
  \texttt{lzy@motern.com} \\
}

\begin{document}
\maketitle

\begin{abstract}
Segment Anything Model 2 (SAM2) demonstrates exceptional performance in video segmentation and refinement of segmentation results.
We anticipate that it can further evolve to achieve higher levels of automation for practical applications.
Building upon SAM2, we conducted a series of practices that ultimately led to the development of a fully automated pipeline, termed Det-SAM2,  in which object prompts are automatically generated by a detection model to facilitate inference and refinement by SAM2.
This pipeline enables inference on infinitely long video streams with constant VRAM and RAM usage, all while preserving the same efficiency and accuracy as the original SAM2.

This technical report focuses on the construction of the overall Det-SAM2 framework and the subsequent engineering optimization applied to SAM2.
We present a case demonstrating an application built on the Det-SAM2 framework: AI refereeing in a billiards scenario, derived from our business context.
The project at \url{https://github.com/motern88/Det-SAM2}.
\end{abstract}

% keywords can be removed

\section{Introduction}
Segment Anything Model 2 (SAM2)\cite{sam2} is currently the state-of-the-art (SOTA) model in video segmentation.
It exhibits advanced object-level instance segmentation capabilities and continues the mask fuzzy matching and interactive refinement features from Segment Anything (SAM)\cite{sam}.
However, the official implementation of SAM2 necessitates user interaction with the initial frame of the video and the application of condition prompts before inference can commence.
During the inference process, if corrections are required for any inaccuracies in SAM2's outputs, users must introduce new condition prompts adjacent to the erroneous frames and re-initiate the inference process.
On one hand, frequent manual interactions constrains the applicability of SAM2 in automated scenarios.
On the other hand, SAM2 requires  a complete re-inference each time new condition prompts or categories are introduced, which significantly exacerbates the already substantial performance overhead associated with SAM2.

To address these challenges, we developed Det-SAM2.
A video object segmentation pipeline that automatically generates prompts for SAM2 using the YOLOv8\cite{yolov8} detection model,This pipeline includes a post-processing step for SAM2's segmentation results, facilitating business decisions in specialized scenarios, all without human intervention.
Det-SAM2 retains the robust object segmentation and refinement capabilities of SAM2 while eliminating the need for manual input of condition prompts, thereby enabling fully automated inference.
Furthermore, we have implemented a series of engineering enhancements to the Det-SAM2 framework, which effectively reduce performance overhead.

Specifically, our key contributions include:
\begin{enumerate}
\item We have developed a self-prompting video instance segmentation pipeline (Det-SAM2-pipeline) that requires no manual interaction for prompt input. It supports inference and segmentation of specific categories, as determined by a custom detection model, using \textbf{video streams}. It returns segmentation results with accuracy comparable to that of SAM2, thereby facilitating post-processing for business applications.
\item We have incorporated the capability to \textbf{add new objects online} during the SAM2 inference tracking process without interrupting the inference state.
\item Our pipeline introduces the application of a memory bank from one video to a new video, referred to as a \textbf{preload memory bank}. This feature allows the system to leverage memories(object categories, shapes, motion states) from the inference analysis of the previous video to assist in conducting similar inference in a new video, \textbf{without the need to add any prompts} in the new video.
\item We have ensured \textbf{constant GPU and memory usage} in the Det-SAM2 pipeline, enabling the inference of infinitely long videos without interruption.
\end{enumerate}

\begin{figure}
  \centering
  \includegraphics[width=0.8\textwidth]{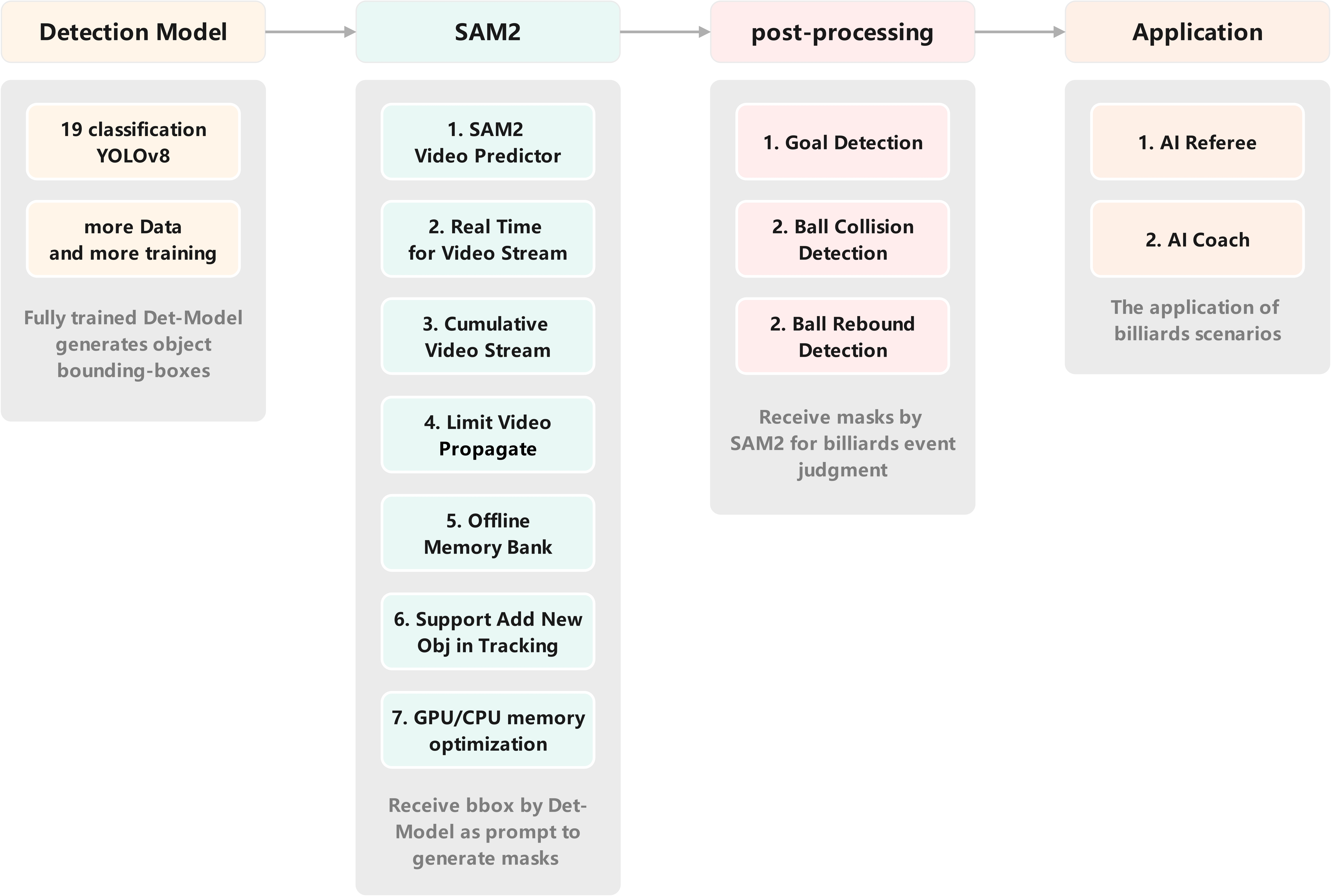}
  \caption{Overview of Det-SAM2 Tasks.
  The overall technical pipeline of Det-SAM2 comprises three key components: the detection module, the pixel-level video tracking module using SAM2 instances, and the post-processing module.
  The detection model provides initial (potentially imperfect) bounding boxes, which are used as conditional prompts for SAM2.
  The SAM2 video predictor propagates these discrete frame prompts (\texttt{propagate\_in\_video}) across all frames in the video, enabling continuous inference.
  Ultimately, the SAM2 video predictor outputs spatiotemporal masks for object instances throughout the video.
  The post-processing module then analyzes the obtained masks to deliver accurate and quantifiable results, thereby supporting higher-level applications such as an AI coach or AI referee in billiards scenarios.}
  \label{fig:fig1}
\end{figure}

Our work focuses exclusively on engineering optimizations and does not involve training or fine-tuning the SAM2\cite{sam2} model itself.
The tasks associated with the implementation of the Det-SAM2 framework and the development of its application example(AI referee for the billiards scene) are illustrated in Figure \ref{fig:fig1}.

\section{Related Work}
\begin{figure}
  \centering
  \includegraphics[width=1.0\textwidth]{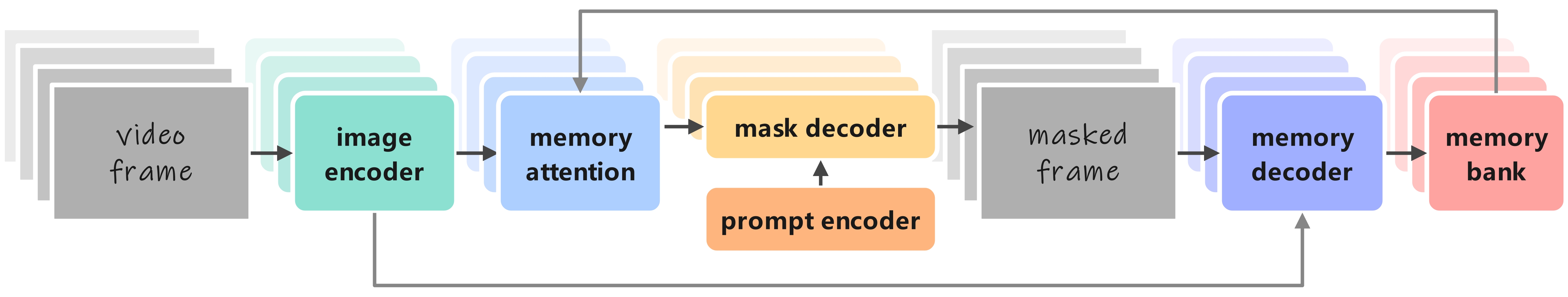}
  \caption{Original Framework of SAM2. The video frame features are processed through Memory Attention, integrating information from the current frame with that in the Memory Bank, and then passed to the Mask Decoder, which uses the conditional prompts to generate the predicted masks.
  The Memory Bank is extracted by the Memory Decoder from the conditional frames.
  The Memory Decoder receives outputs not only from the Mask Decoder but also from the Image Encoder.}
  \label{fig:fig2}
\end{figure}

\textbf{Segment Anything Model 2}\cite{sam2} achieves object segmentation in videos by leveraging its comprehensive understanding of object-level concepts.
It effectively manages deformations such as stretching and occlusion, representing a significant advancement beyond traditional detection and segmentation models.
Furthermore, SAM2 exhibits robust correction capabilities, allowing newly provided conditional prompts to be applied retroactively across all previously inferred frames, thereby facilitating the correction of specific errors.
The SAM2 Video Predictor, as illustrated in Figure \ref{fig:fig2}, maintains a Memory Bank that stores the input image features and predicted mask results for each frame.
During the prediction of the segmentation mask for each frame, the inference computation incorporates the Memory Bank, conditional prompts, and input image features.

\textbf{YOLOv8}\cite{yolov8} is a new version of the YOLO series launched by Ultralytics in 2023.
It is built upon YOLOv5\cite{yolov5}, incorporating significant architectural and methodological innovations.
With its outstanding performance and broad applicability, YOLOv8 has emerged as the preferred choice for industrial deployment in the field of object detection.

\section{Methodology}
In this section, we focus on how we built the Det-SAM2 pipeline corresponding to the seven subtasks illustrated in Figure \ref{fig:fig1}, as well as the series of engineering challenges encountered and addressed throughout this process.

\subsection{Detection Model + SAM2 Video Predictior}
As previously mentioned, interactive prompts are essential for SAM2's\cite{sam2} accurate segmentation abut also pose a barrier to achieving fully automated inference without human intervention.
The initial prompt for SAM2's first input frame must be manually provided to start the inference process.
We explored the possibility of utilizing a detection model to eliminate the need for human-provided prompts.

\begin{figure}[H]
  \centering
  \includegraphics[width=1.0\textwidth]{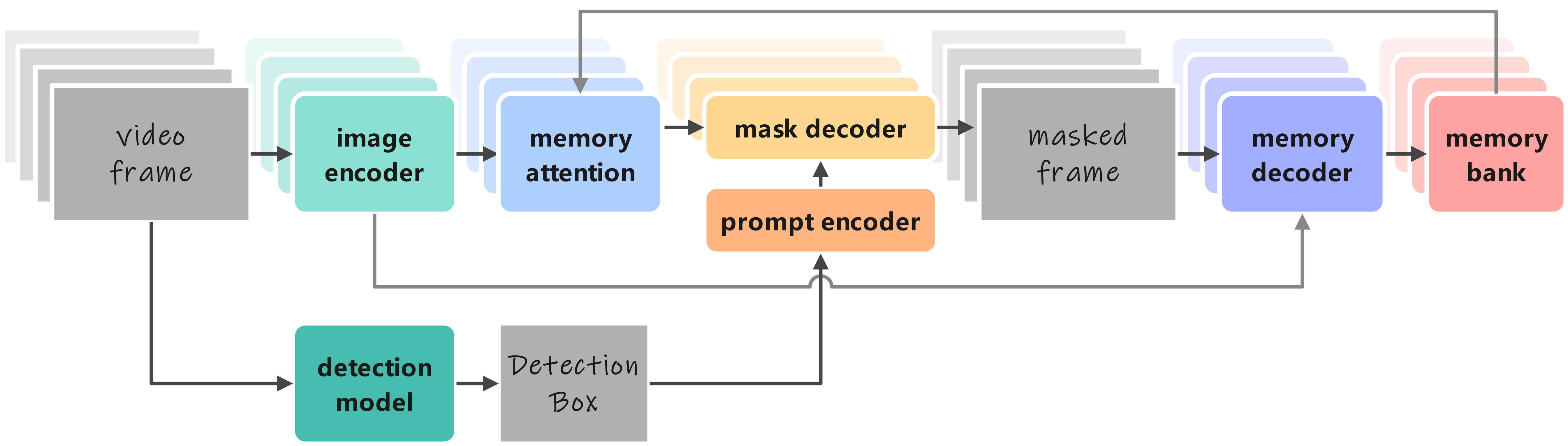}
  \caption{Det-SAM2 Experimental Demo Framework Diagram.
  The condition prompt for a given frame is automatically added, and the condition prompt is provided by the detection model (in this case, YOLOv8).
  The detection box results are used as the prompt input for the Prompt Encoder.}
  \label{fig:fig3}
\end{figure}

As shown in Figure \ref{fig:fig3}, compared to Figure \ref{fig:fig2}, we introduced a branch that connects the video frame input to the prompt input.
This branch is powered by a detection model, which applies prompts in the form of detection boxes to the video frames for our specified categories.
With this addition, SAM2 can commence predictions without human intervention.
This marks the initial form of Det-SAM2 (Detection Model + SAM2 Video Predictor).

\subsection{Det-SAM2 in Video Stream}
Once our Det-SAM2 system is capable of initiating segmentation predictions without manual intervention, we also seek it to leverage SAM2's powerful correction capability.
Rather than relying solely on the prompt information from the initial frame to infer the predictions for the entire video.
We aim to continuously add box prompts automatically generated by the detection model to SAM2 in the video stream.
Due to SAM2's correction mechanism, whenever SAM2 receives a new prompt, it propagates the new prompt information to all previously inferred frames (using \texttt{propagate\_in\_video}).
Consequently, the memory bank, now updated with the newly received prompt, is reintegrated into the memory attention of each previously processed frame, allowing for the recalculation of segmentation masks and enabling effective corrections.

\begin{figure}
  \centering
  \includegraphics[width=1.0\textwidth]{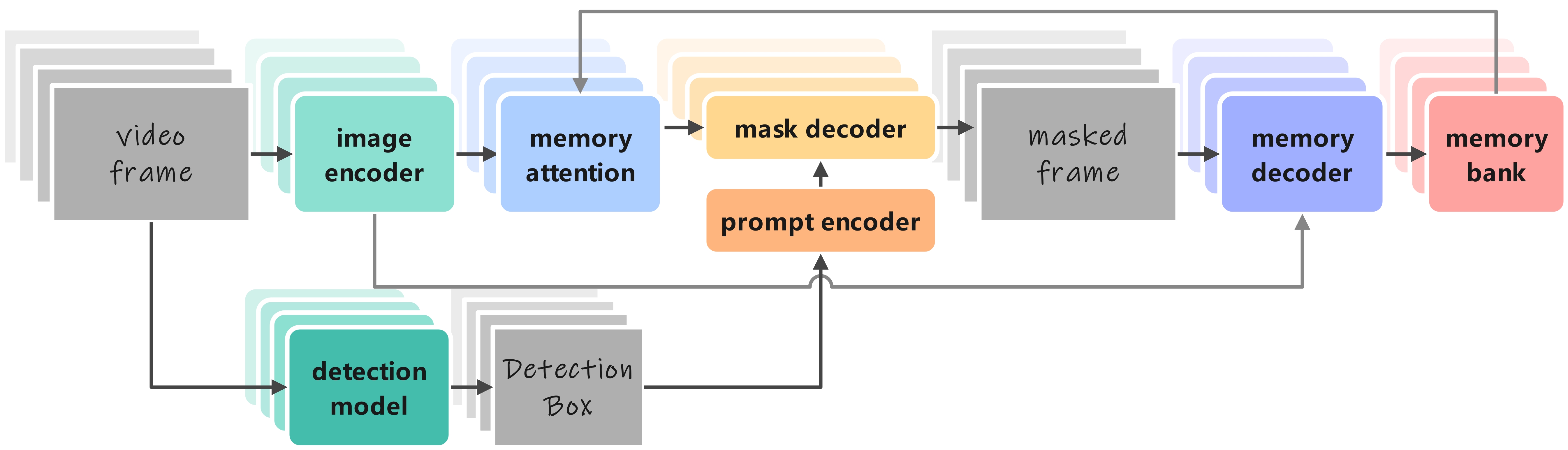}
  \caption{The Det-SAM2 framework  facilitates for the automatic addition of condition prompts for each frame.
  In contrast to Figure \ref{fig:fig3}, where the detection model branch is not only active in the initial frame of the video, it is now applied to every frame throughout the video.}
  \label{fig:fig4}
\end{figure}

\begin{figure}
  \centering
  \includegraphics[width=0.65\textwidth]{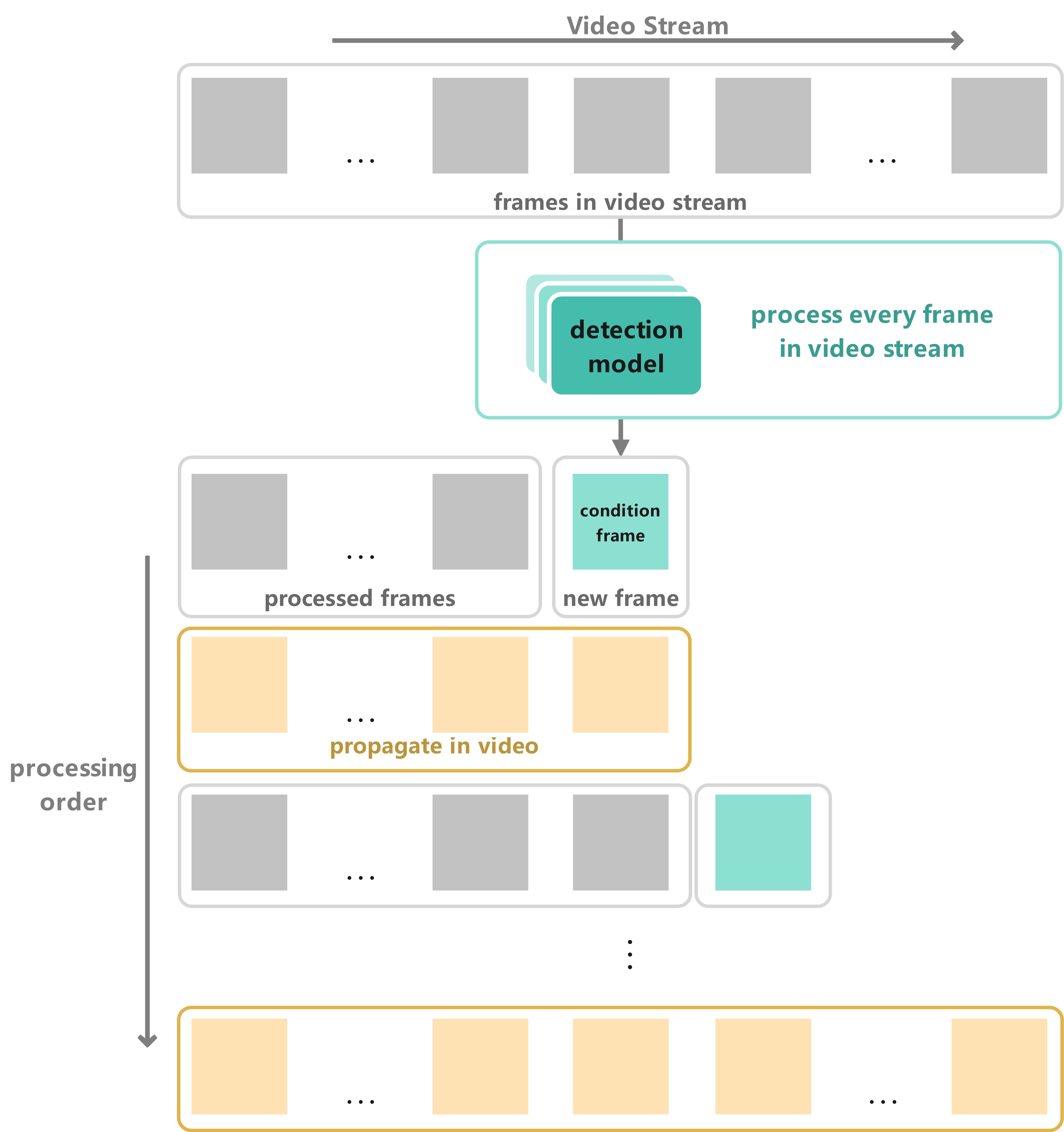}
  \caption{Det-SAM2 Video Stream Processing Diagram.
  Each frame passes through the Detection Model as a condition frame for SAM2 (represented in green in the diagram), and then the propagation operation (depicted in yellow as "propagate in video") is applied to all previously processed video frames to enable the correction capability.}
  \label{fig:fig5}
\end{figure}

The implement the automatic addition of condition prompts for each frame within the Det-SAM2 framework is illustrated in Figure \ref{fig:fig4}.
The primary distinction between this figure and Figure \ref{fig:fig3} is that we invoke the detection model branch for each frame.
Additionally, the schematic representation of Det-SAM2 processing the video stream over time is depicted in Figure \ref{fig:fig5}.

However, as shown in Figure \ref{fig:fig5}, in practice, the propagation process (\texttt{propagate\_in\_video}) consumes a considerable amount of inference time.
This is because, during the propagation, each time a new condition frame is added, SAM2 re-infers all previous frames.
As a result, when processing a video of length $N$ frames, the current framework requires inference for a total of $ \frac{1}{2} N^2 $ frames.

\subsection{Cumulative Video Stream}
A straightforward method to optimize inference overhead is to minimize the frequency of inference propagation (\texttt{propagate\_in\_video}).
This reduction can be approached from two perspectives: One approach is to reduce the frequency at which condition frames are generated, as only these frames trigger SAM2's correction mechanism, leading to repeated inference propagation on previously processed frames.
Another approach is to increase the number of video frames in a single input. When multiple frames are input simultaneously, SAM2 performs propagation only once.

To implement this, we aim to space out the inference of the detection model so that not every frame input into SAM2 requires to have condition prompts from the detection model.
Additionally, We propose allowing new incoming video frames to accumulate in a video frame buffer while receiving the video stream.
Instead of inputting each frame into the Det-SAM2 framework individually, we input a sequence of accumulated frames all at once.
This way, SAM2 processes multiple frames in a single pass, thereby reducing the number of propagation steps.

\begin{figure}
  \centering
  \includegraphics[width=0.8\textwidth]{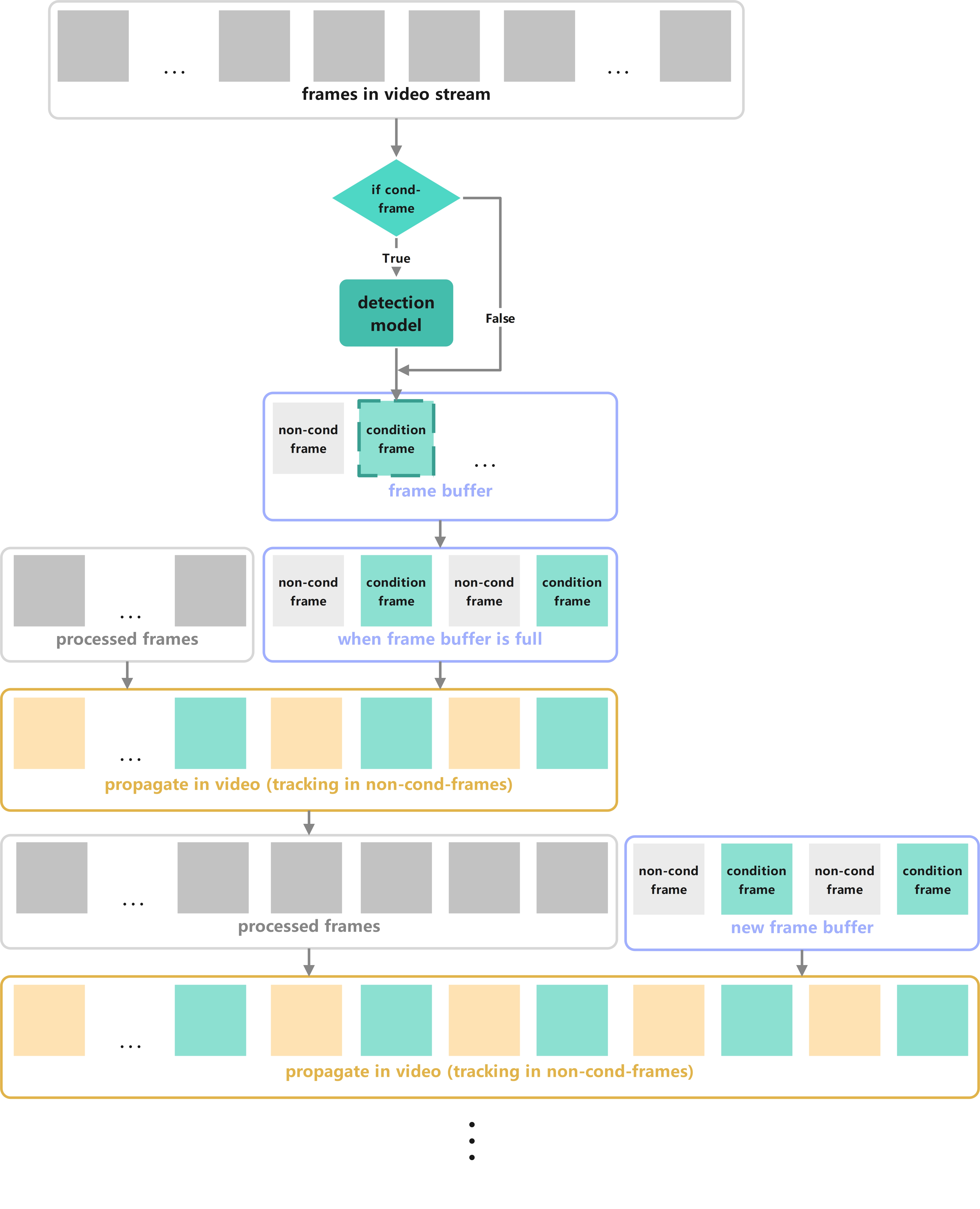}
  \caption{Flowchart of cumulative video stream and interval-based detection condition prompts in Det-SAM2.
  For each frame received in the video stream, it will first be accumulated in the frame buffer. When the frame buffer accumulates enough frames, we will input the sequence of frames from the buffer into the Det-SAM2 framework all at once.
  During inference of the current video frame sequence, SAM2 will determine which frames are condition frames and which are non-condition frames based on the interval setting.
  Condition frames are provided with condition prompts by the detection model, and the prompt embeddings are generated by SAM2's Prompt Encoder.}
  \label{fig:fig6}
\end{figure}

The diagram illustrating the process of accumulating the video stream and spacing out the condition prompts from the detection model is shown in Figure \ref{fig:fig6}.

By accumulating a certain number of video frames before inference, we can significantly reduce the number of propagation steps in SAM2.
Suppose we accumulate a sequence of $K$ frames at a time.
When performing inference on an entire video of length $N$ frames, the propagation process (\texttt{propagate\_in\_video}) would only need to process approximately $ \frac{1}{2K} N^2 $ frames.

\subsection{Limited Video Propagate}
We already know that SAM2\cite{sam2} achieves prompt-based corrections by receiving new conditional prompts and performing re-inference propagation(\texttt{propagate\_in\_video}) across all historical frames.
In the case of a static video (i.e., when the entire video is input at once), it is understandable that each propagation must be executed across all frames.
However, during inference on a video stream, it is unnecessary for the propagation process to be applied to all historical frames.

In the actual Det-SAM2 process, the non-condition frames that require correction are typically not far removed from the condition frames they depend on.
Therefore, we can limit the number of frames involved in the inference during each propagation.
During each inference process in the video stream,most distant past frames are already finalized and do not require correction.
In contrast, the more recent a past frame is, the more likely its inference results may be altered in future predictions.
Therefore, we impose the following restrictions on the propagation(\texttt{propagate\_in\_video}):
\begin{enumerate}
\item Set the \texttt{propagate\_in\_video} to process frames in reverse order, starting from the most recent frame.
\item Limit the maximum number of frames to be processed during propagation(\texttt{propagate\_in\_video}). However, the length of the propagation should at least encompass the cumulative video frame sequence from the previous inference; otherwise, the propagation would lose its corrective significance.
\end{enumerate}

\begin{figure}
  \centering
  \includegraphics[width=1.0\textwidth]{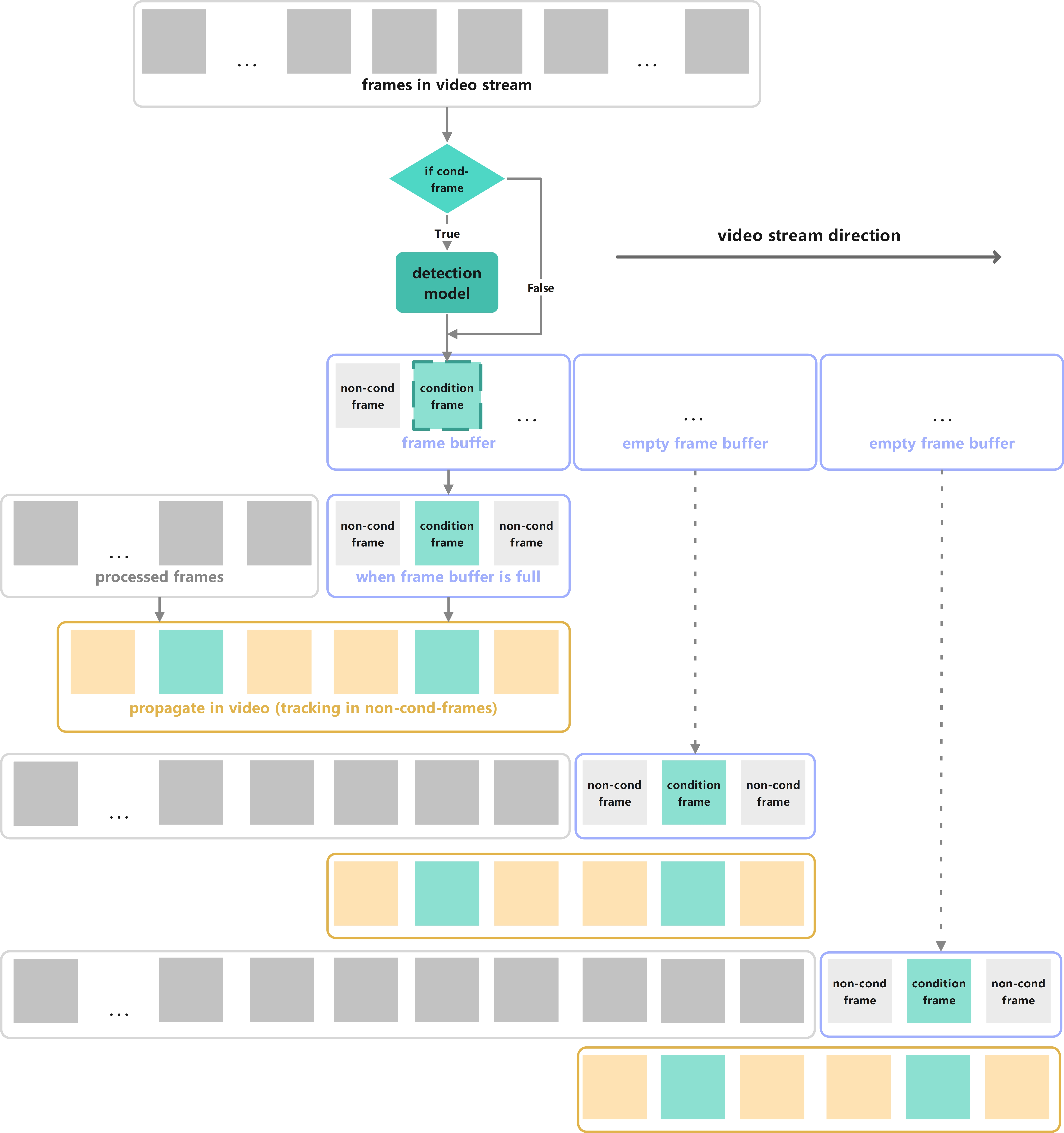}
  \caption{Flow Diagram of Limited Video Propagation: During propagation (propagate in video), the process is restricted to a user-defined maximum tracking length (\texttt{max\_frame\_num\_to\_track}), instead of iterating through all previous frames.}
  \label{fig:fig7}
\end{figure}

The process diagram for Limited Video Propagation is illustrated in Figure \ref{fig:fig7}.

Increasing the maximum propagation length (\texttt{max\_frame\_num\_to\_track}) can expand the correction range but will result in higher computational overhead.
Conversely, reducing the maximum propagation length can accelerate inference speed but will limit the range within which conditional frames can correct non-conditional frames in the video stream.

By limiting the propagation length(\texttt{max\_frame\_num\_to\_track}) to $M$, and the cumulative frame buffer size (\texttt{max\_frame\_num\_to\_track}) to $K$, inference for a video of length $N$ frames requires processing approximately $ \frac{M}{K} N $ frames.
This implies that when $M$ and $N$ are fixed, the actual number of frames processed by Det-SAM2 increases linearly with the video length.

\subsection{Preload Memory Bank(Offline Memory Bank)}
In comparison to SAM\cite{sam}, a significant enhancement in SAM2\cite{sam2} that facilitates the transfer of image segmentation capabilities to video segmentation is the incorporation of the Memory Bank.
This Memory Bank establishes associations between frames through Memory Attention, allowing SAM2 to leverage temporal context across video frames for improved segmentation performance.
However, the generation and construction of the Memory Bank must be online, meaning it needs to be built in real-time during the inference process.
Each time a new video frame is input, the Memory Bank is updated with the condition prompt, output mask, and video frame feature through the Memory Decoder.

\begin{figure}
  \centering
  \includegraphics[width=1.0\textwidth]{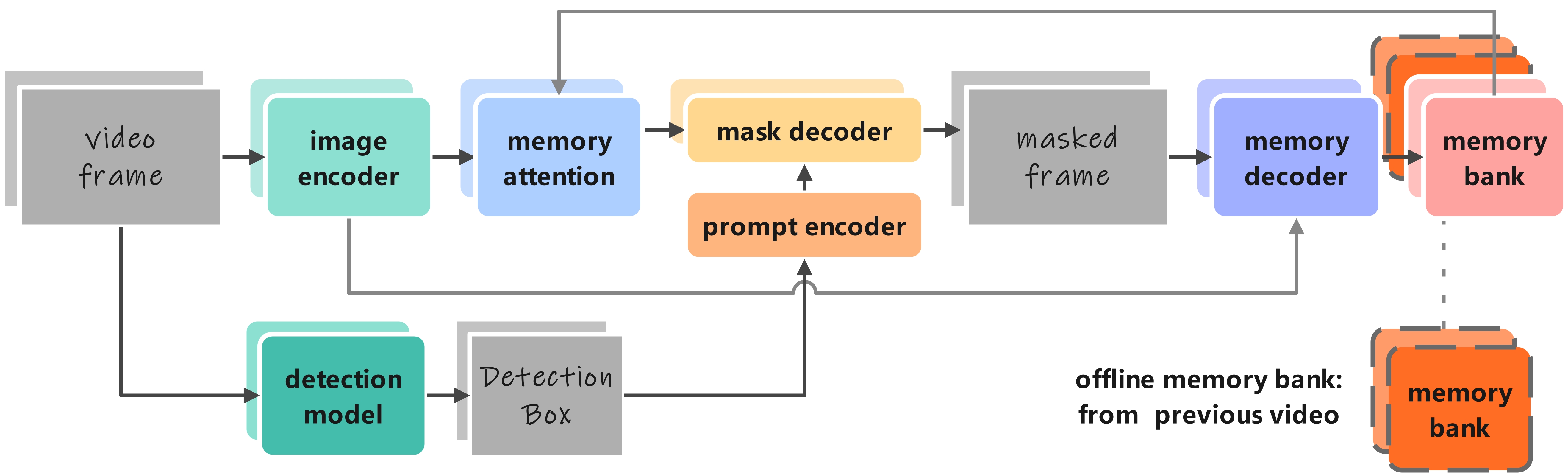}
  \caption{Preload memory bank flowchart. The Memory Bank preloads an offline memory bank, sourced from previously inferred videos.
  As a result, during the current inference, the previously inferred memory from past videos can be directly applied to the inference of the new video.}
  \label{fig:fig8}
\end{figure}

Inspired by discussions in the official code repository issues\cite{issue210}, we propose preload a Memory Bank that has already been constructed from an old video.
This approach enables the utilization of memory information from the previous video for the new video, allowing inference without the need to add any condition prompts.
This entails allowing SAM2 to preload an offline Memory Bank that has been meticulously designed to include all necessary prompts and challenging sample prompts that may be required for the new video, similar to a "system prompt".
During subsequent inference on the new video, the newly generated memory will be accumulated on top of the preloaded Memory Bank.
Although the newly generated memory and the preloaded memory are conceptually distinct, there is no practical differentiation between the two during actual inference in SAM2.
The process of preloading the memory bank is illustrated in Figure \ref{fig:fig8}.

In the specific implementation, the \texttt{inference\_state} in SAM2 holds all the information in the memory bank.
Therefore, by migrating the \texttt{inference\_state} to the new video inference, the new video can directly leverage the existing memory without needing to reinitialize the \texttt{init\_state}.

\subsection{Support Add New Objects in Tracking}
Currently, in the official SAM2 implementation, it is only permissible to predefine the instances that need to be segmented before tracking begins, and it does not support the online addition of new instances during the tracking process.
The specific reason is that when the memory bank is first initialized (\texttt{init\_state}), the class mapping list is fixed.
If new object categories that need to be inferred are introduced in a new frame, it results in a mismatch in the feature tensor sizes between the new frame and the old frames, making it impossible to compute.

n practice, during the real-time video stream processing with Det-SAM2, we often cannot predict which objects will appear in the future.
Our detection model is likely to output only a partial set of categories when initializing the memory bank for the first time.
Consequently, any remaining categories that are introduced to SAM2 after inference has commenced will be treated as new objects requiring processing.
This situation is unavoidable.
The official solution to this is to perform a \texttt{init\_state}, which resets the memory bank.
This means that in the frame where a new category appears, the memory bank is reinitialized. The consequence of this approach is that all previous inference results are lost, and segmentation inference starts anew from that point.

Moreover, performing a reinitialization each time a new category appears in a frame cannot guarantee that the reinitialized frame sequence will contain all potential categories.
Therefore, any method that involves resetting the memory bank upon encountering a new category will inevitably trigger a \texttt{reset\_state} initialization repeatedly in long videos.
This continuous resetting to avoid tensor size mismatch errors between frames due to different numbers of categories, and the subsequent loss of all previous inference results, effectively clears the memory bank each time, rendering the process inefficient and preventing the model from leveraging prior context across extended video sequences.

To address this issue, we propose enabling the natural addition of new object IDs during the tracking process without resetting the entire memory bank.
We achieve this by updating the memory bank online during the tracking process.
To enable the inference to naturally add new object categories during the tracking process, when encountering a new object category, we need to update the following operations:
\begin{enumerate}
\item Register a new ID's corresponding index list and information storage dictionary in the \texttt{inference\_state}.
\item Update the memory bank of all previous frames using the new ID mapping table (reacquire \texttt{output\_dict} or \texttt{temp\_output\_dict}, and generate the memory bank under the new ID mapping relationships through the Memory Encoder).
\end{enumerate}

\begin{figure}[H]
  \centering
  \includegraphics[width=1.0\textwidth]{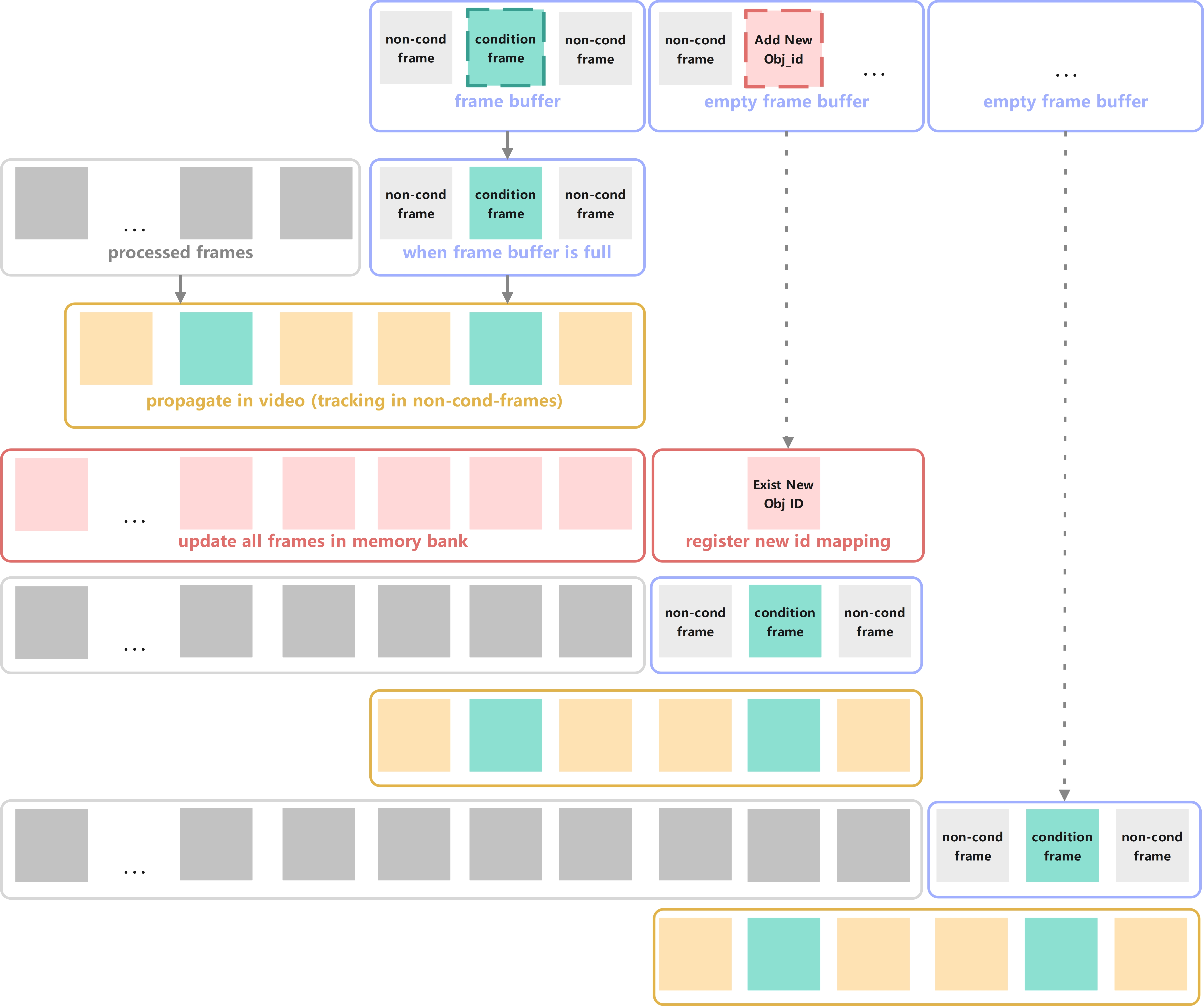}
  \caption{The schematic diagram of online memory bank updates when Det-SAM2 adds new objects during tracking.
  When the framework receives new object categories after initiating inference and tracking, it first registers the new object IDs in the memory bank.
  Subsequently, based on the updated ID mapping table, the memory bank for all previous historical frames is updated.
   (The batch dimension size of the information tensors in the memory bank for each frame depends on the number of IDs to be predicted. Without updating, tensor mismatches will occur, preventing the calculation of Memory Attention.)}
  \label{fig:fig9}
\end{figure}

The schematic diagram of online updating of the memory bank during tracking is shown in Figure \ref{fig:fig9}.

After implementing the functionality shown in Figure \ref{fig:fig9}, we now support the ability to add new object IDs online after tracking has started.
However, in long video inference,  the current method for online updating of the memory bank processes all historical frames in the memory bank using the memory encoder each time it is applied.
If a new object appears at the end of a long video, this approach incurs a significant computational overhead.

\begin{figure}[H]
  \centering
  \includegraphics[width=1.0\textwidth]{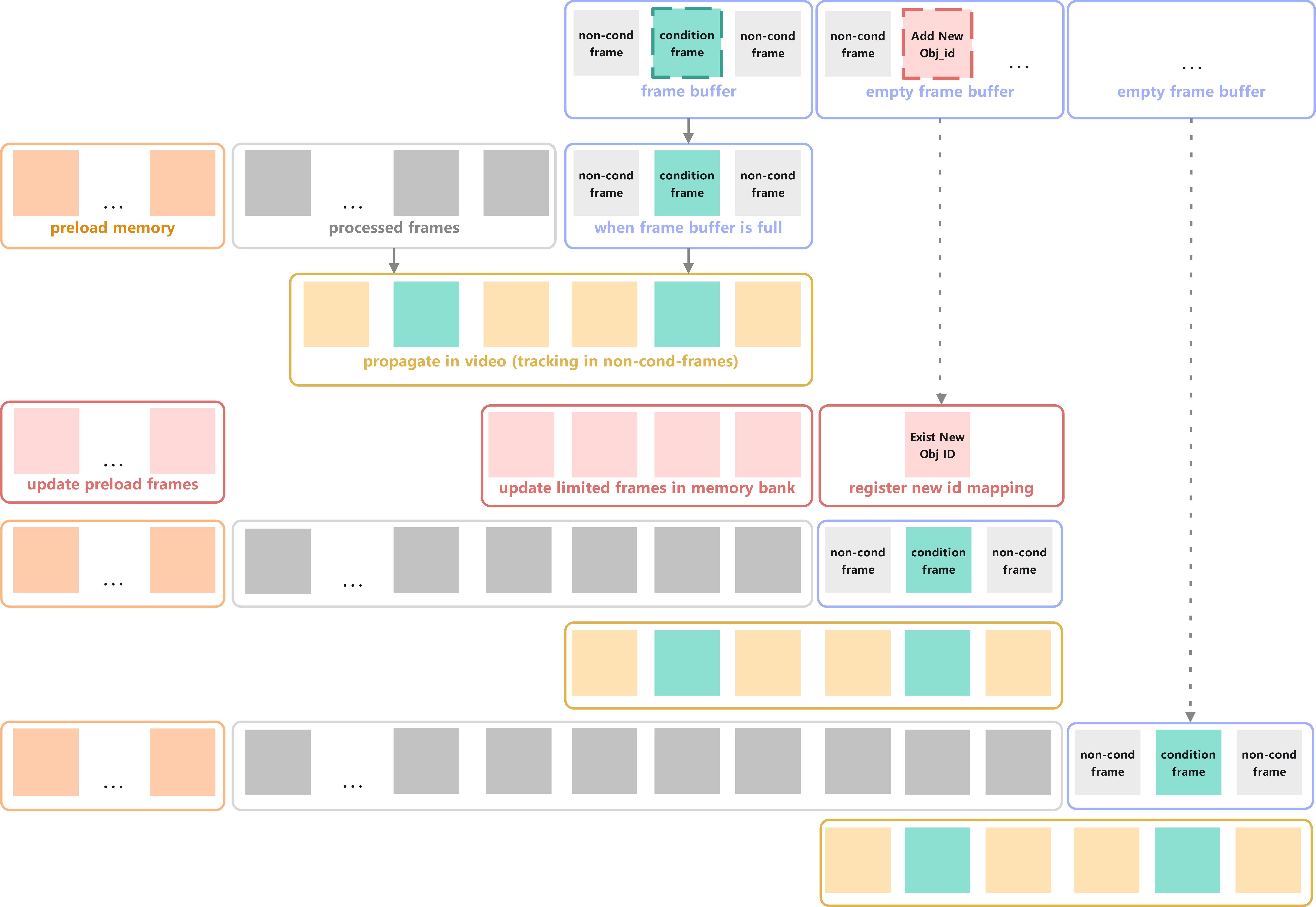}
  \caption{The optimized schematic diagram for online memory bank updates when new object IDs are input during the Det-SAM2 tracking process.
  By limiting the maximum number of frames updated in the Memory Bank and the number of condition frames used in Memory Attention calculation, while ensuring that the frames in the Preload Memory Bank are updated and involved in Memory Attention calculation, we can guarantee performance while reducing computational overhead.}
  \label{fig:fig10}
\end{figure}

To improve performance efficiency, we need to impose restrictions on two components:
\begin{enumerate}
\item Limit the number of frames updated in the memory bank when a new object category appears.
\item Limit the number of frames used for memory attention calculation to avoid using old frames that have not been updated.
\end{enumerate}
The specific approach is as follows:
\begin{enumerate}
\item When a new category is added during inference and tracking, only a limited number of frames close to the current moment in the memory bank are updated, ensuring that all frames in the preload memory bank are also updated.
\item Restrict the number of frames used for memory attention calculation while ensuring that all condition frames in the preloaded memory bank are included in the computation.
\end{enumerate}
The optimized process is shown in Figure \ref{fig:fig10}.

\subsection{GPU/CPU memory optimization}
The official source code for SAM2\cite{sam2} supports running on CPU, GPU, and NPU.
However, since our own devices primarily use CPU and GPU, our optimizations mainly focus on GPU memory (VRAM) and system memory (RAM).
In this section, we address the reduction of RAM and VRAM usage during the inference process in the Det-SAM2 architecture.
We will introduce several interfaces reserved in the official source code, during which we will refer to many function names from the SAM2 source code.

Before optimization, the Det-SAM2 framework can infer approximately 200 video frames per 24GB of VRAM, with 6-7 segmented objects per frame.
At this point, the RAM and VRAM usage increases linearly with the total number of frames in the video being processed.

\begin{enumerate}
    \item First Optimization Interface: We will explore the first optimization interface reserved by the official source code, which is the \texttt{offload\_video\_to\_cpu} parameter in the \texttt{SAM2VideoPredictor.init\_state()} method. This parameter allows transferring the video frames in the memory bank (i.e., \texttt{inference\_state["images"]}) from GPU VRAM to CPU RAM. At a video resolution of 1920x1080, this can reduce approximately 0.025GB of VRAM usage per frame, which means a reduction of 2.5GB of usage for every 100 frames.

    \item Second Optimization Interface: Next, we will try the second optimization interface reserved by the official source code, which is the \texttt{offload\_state\_to\_cpu} parameter in the   \texttt{SAM2VideoPredictor.init\_state()} method. The purpose of this parameter is to store large feature tensors, which do not require frequent computation, in CPU memory. However, in the Det-SAM2 framework we built, using this parameter does not directly save VRAM. Instead, it causes a misalignment between the generated segmentation masks and the frame indices.

    It was only after we set the tensor transfer parameter \texttt{non\_blocking=False} at all locations where the \texttt{inference["storage\_device"]} tensor that this interface began to function correctly.
    \begin{verbatim}
    device = inference_state["storage_device"]
    tensor.to(device,non_blocking=False)
    \end{verbatim}
    When \texttt{offload\_state\_to\_cpu=True}, the final effect after the fix, as mentioned in the official comments, is that VRAM usage is reduced while the inference time increases by approximately 22\%.

    \item Inspired by the discussion in the official repository issue\cite{issue196-1}, we hope to try clearing old frame data continuously (under the condition that the old frame data will no longer be used) in order to prevent the total memory usage from increasing indefinitely. To implement this functionality, we added the \texttt{release\_old\_frames()} method in the \texttt{SAM2VideoPredictor} class within \texttt{sam2.sam2\_video\_predictor}. This method allows setting a maximum number of frames to retain (\texttt{max\_inference\_state\_frames}). Frames that are older than the maximum retention distance from the current frame will be considered as frames to be cleared.

    Therefore, to ensure that only frames that will no longer be used are cleared, \texttt{max\_inference\_state\_frames} should be greater than the maximum propagation length (\texttt{max\_frame\_num\_to\_track}) in \texttt{propagate\_in\_video()}.

    The schematic diagram of the continuous old frame clearing process is shown in Figure \ref{fig:fig11}, and the implementation of \texttt{release\_old\_frames()} can be found in Appendix \ref{A1}. With this approach, constant VRAM usage can be maintained during the inference process of infinitely long video frames.

    \begin{figure}
        \centering
        \includegraphics[width=0.9\textwidth]{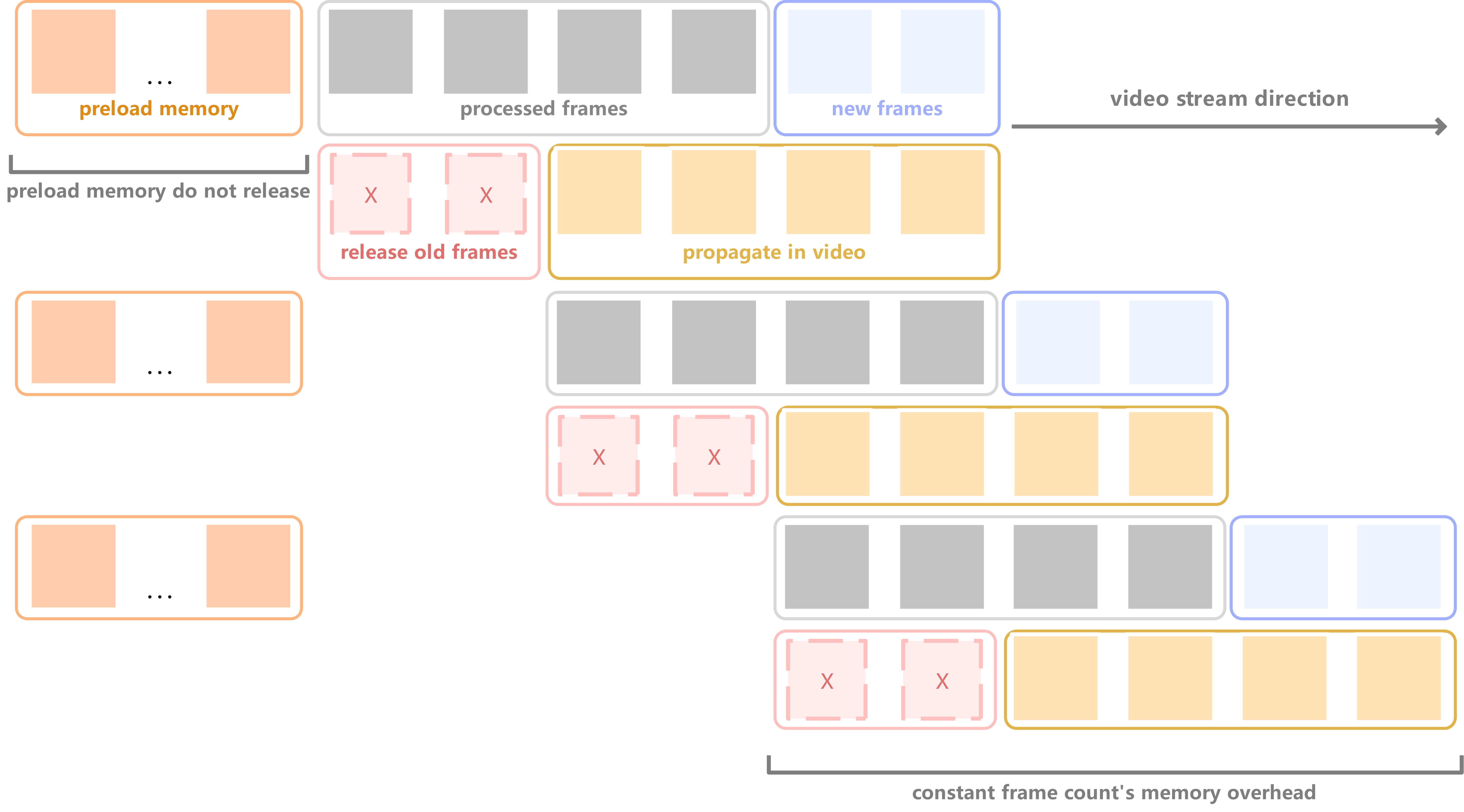}
        \caption{Schematic diagram of continuously releasing old frames to maintain constant VRAM usage.}
        \label{fig:fig11}
    \end{figure}

    The diagram Figure \ref{fig:fig11} illustrates the case where the maximum number of retained frames is equal to the maximum propagation length (in the example, \texttt{max\_inference\_state\_frames = max\_frame\_num\_to\_track = 4}). After each propagation (\texttt{propagate\_in\_video}), any processed frames that exceed the maximum retention limit (4 frames) will be released and cleared. In addition, the frames in the preload memory bank should never be released. Therefore, in this example (where the maximum number of retained frames equals the maximum propagation length) during the stable inference process of a long video:
    \begin{enumerate}
        \item The upper limit of VRAM usage is the data usage of \texttt{len(new\_frames) + max\_inference\_state\_frames + len(preload\_memory)} frames.
        
        \item The lower limit of VRAM usage is the data usage of \texttt{max\_frame\_num\_to\_track + len(preload\_memory)} frames.
    \end{enumerate}

    \item We unexpectedly found that the Memory Attention calculation in our pipeline generates a large number of intermediate variables that occupy VRAM and are not released in a timely manner. Therefore, we need to manually release the VRAM after the Memory Attention calculation. We discovered that this can significantly reduce the upper limit of VRAM usage in the Det-SAM2 pipeline. The specific implementation details can be found in Appendix \ref{A2}.

    \item Similarly, inspired by issue\cite{issue196-2} in the official code repository, we tried storing the images in FP16 half precision instead of the original FP32. This saves approximately 0.007GB/frame of memory at a resolution of 1920x1080, with almost no loss in segmentation mask quality.

    \item To further reduce the linear growth of memory usage while keeping VRAM usage constant, we aim to continuously clear old data, including clearing the cached video frames in \texttt{inference["images"]}. Although these frames have already been offloaded to CPU memory using the   \texttt{offload\_video\_to\_cpu=True} parameter, we still want to maintain a constant memory usage for this part (at a resolution of 1920x1080, 0.025GB/frame of memory usage grows linearly).

    To implement this functionality, we need to make four changes:
    \begin{enumerate}
        \item Modify the \texttt{SAM2VideoPredictor.\_get\_image\_feature()} method, which originally directly retrieves image frames from the corresponding frame index, to instead rely on an independent index mapping list (used to record the non-continuous index relationship in the video frame tensor) to fetch the corresponding frame tensor:
        \begin{verbatim}
target_idx = inference_state["images_idx"].index(frame_idx)
image = inference_state["images"][target_idx].to(device).float().unsqueeze(0)
        \end{verbatim}
        
        \item Implement the registration and updating of the \texttt{inference\_state["images\_idx"]} index mapping table in the \texttt{SAM2VideoPredictor.init\_state()} and \texttt{update\_state()} methods, respectively.
        
        \item Add a function to clear old image frames in the \texttt{SAM2VideoPredictor.release\_old\_frames()} method. This will also synchronize the updates of \texttt{inference\_state["images"]} and \texttt{inference\_state["images\_idx"]}.
        
        \item Modify all references to \texttt{inference\_state["num\_frames"]} in the  \texttt{SAM2VideoPredictor} class to ensure it always represents the total number of video frames that have been loaded historically, rather than the current total frame count (which may have deleted  some old video frames).
    \end{enumerate}

    \item We have successfully achieved constant VRAM usage,  there remains a linear increase in memory consumption due to the ongoing clearing of the \texttt{inference["images"]} video frame cache. The question arises: How can we attain constant memory usage? In fact, the persistent linear growth in memory usage is primarily attributable to the segmentation result dictionary, \texttt{video\_segment} that is continuously collected during the inference process:
    \begin{verbatim}
video_segments[out_frame_idx] = {
	out_obj_id: (out_mask_logits[i] > 0.0).cpu().numpy()
	for i, out_obj_id in enumerate(out_obj_ids)
}
    \end{verbatim}
    By simply releasing the corresponding frame in the \texttt{video\_segments} dictionary after processing the segmentation results for each frame in various downstream tasks, we can achieve constant memory usage:
    \begin{verbatim}
import gc
...
# After your post-process in every frames
video_segments.pop(frame_idx, None)
gc.collect()
    \end{verbatim}

\end{enumerate}

\section{Experiments}
We have integrated Det-SAM2 with a post-processing algorithm tailored to our specific business scenario (see Appendix \ref{B1}) to create a Det-SAM2-pipeline (see Appendix \ref{B2}).
we present a visualization of the Det-SAM2 pipeline and provide an explanation of its functionality.

\begin{figure}[H]
  \centering
  \includegraphics[width=0.5\textwidth]{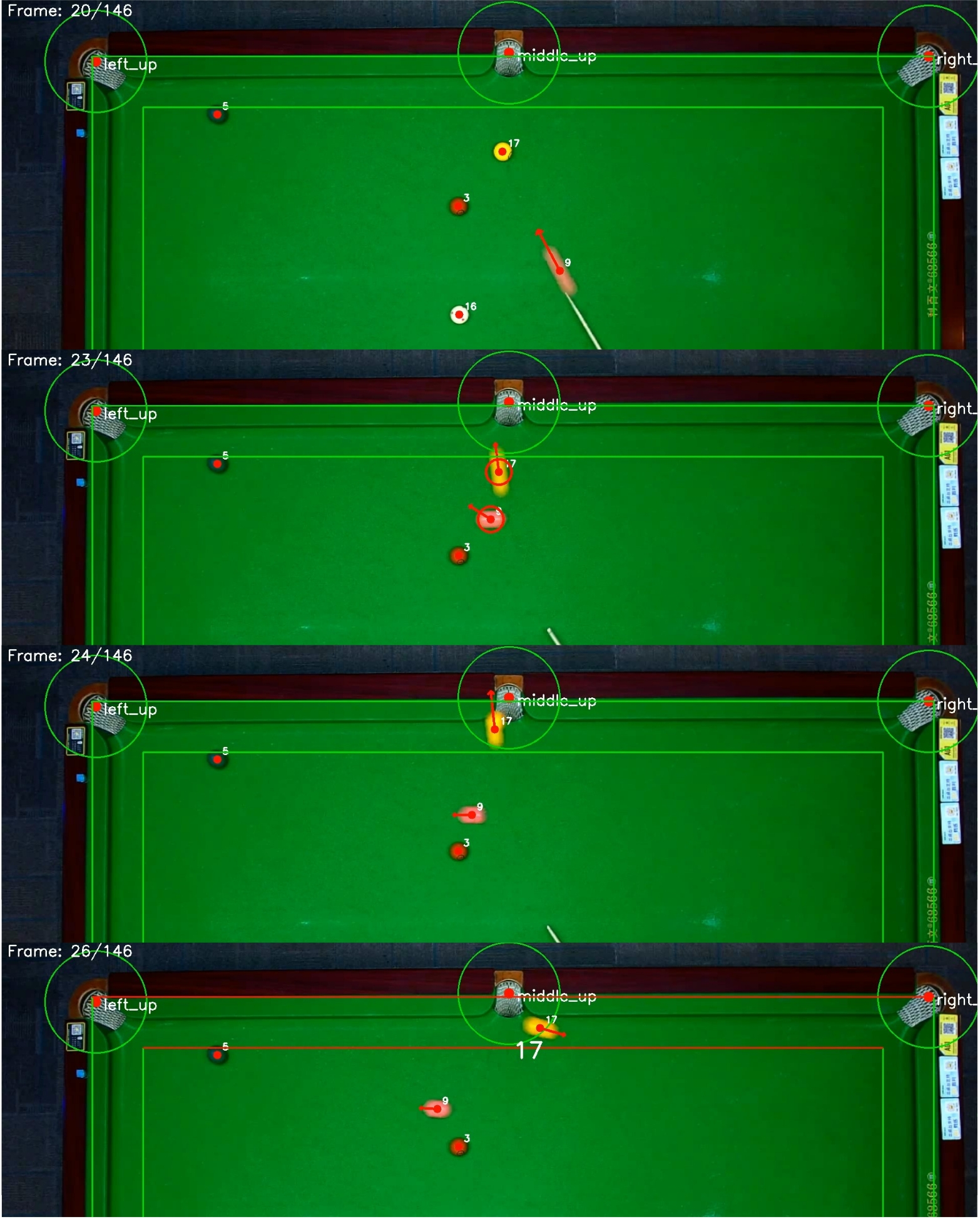}
  \caption{The visualized post-processing result of the Det-SAM2-pipeline in our self-implemented billiard scene example is shown in Figure \ref{fig:fig13}.
  This image was derived from the segmentation masks predicted by SAM2, followed by post-processing.
  As seen, with the support of SAM2, even though the fast-moving balls are stretched in the camera frame, they are still perfectly captured.}
  \label{fig:fig12}
\end{figure}

In the context of high-speed ball movement on a billiard table, Det-SAM2 can automatically infer long videos with the original accuracy of SAM2.
It can accurately track the deformation and stretching of the ball (such as the ball with ID 9 in Figure \ref{fig:fig12}), detect collisions between balls (such as the balls with IDs 9 and 17 in Figure \ref{fig:fig12}), and accurately identifies when a ball bounces off the table's edge (such as the ball with ID 17 in Figure \ref{fig:fig12}).

\begin{figure}
  \centering
  \includegraphics[width=0.5\textwidth]{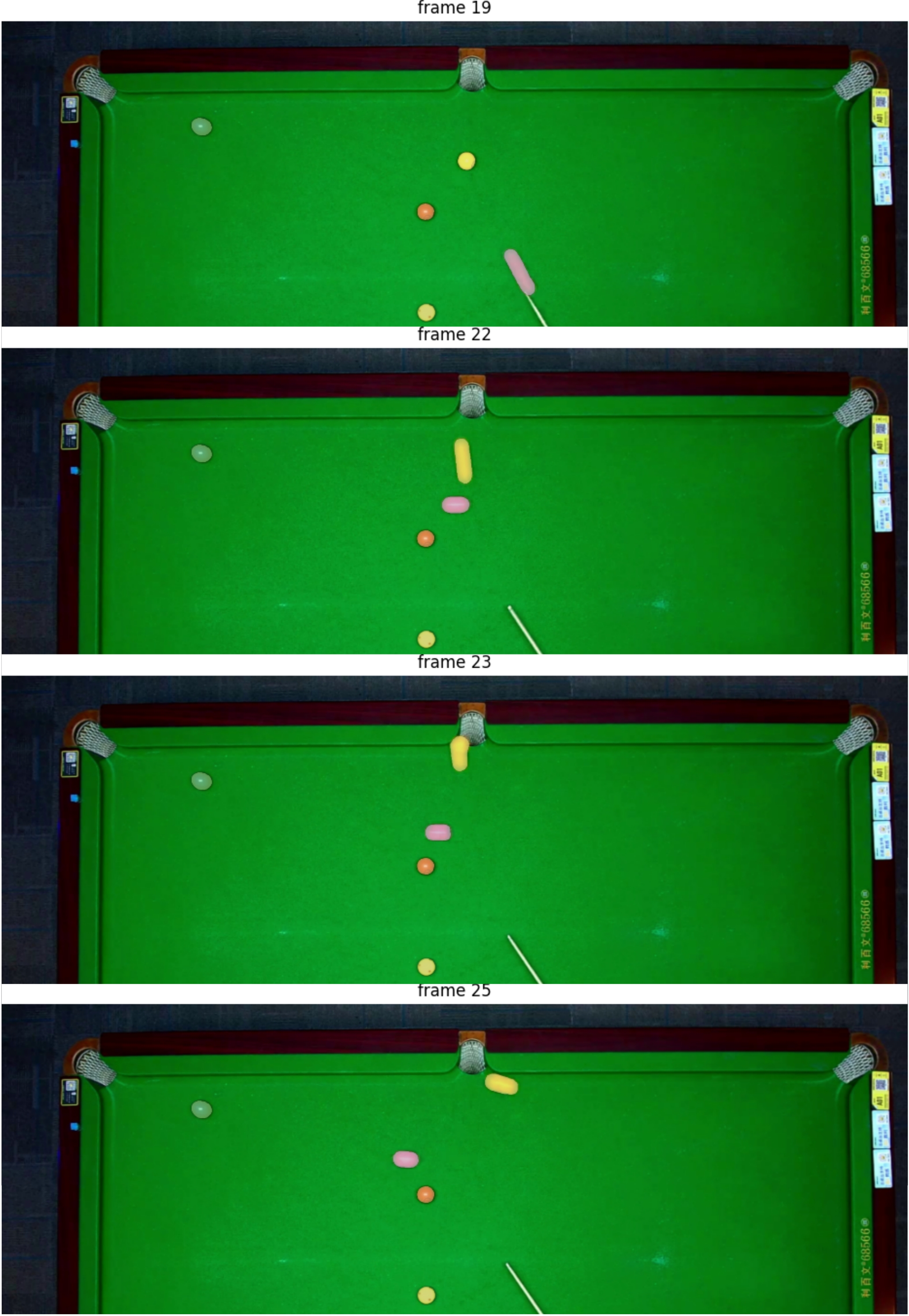}
  \caption{Segmentation mask rendering of the Det-SAM2-pipeline in our self-implemented billiard scene example.
  The Det-SAM2 framework is able to maintain the original segmentation capabilities of SAM2 while operating autonomously.}
  \label{fig:fig13}
\end{figure}

\section{Discussion}
\textbf{1}. During the implementation process, we observed that the frequent generation of conditional prompts for SAM2 within the Det-SAM2 framework leads to an excessive dependence on the detection model.
When certain frames lack prompts, SAM2 fails to predict segmentation results in scenarios where it could have normally performed well, with little to no additional interference.
Therefore, it is necessary to carefully adjust the intervals at which the detection model is activated in custom scenarios to find the optimal parameters.
We have developed an evaluation script specifically designed to assess the best combination of parameters in various example scenarios.

\textbf{2}. In our series of engineering implementations for resource optimization, we have constrained the capacity of the memory bank to store only a limited number of recent frame hints. 
This limitation will inevitably have an impact on videos with a large object association span.
The positive aspect is that our detection branch can consistently provide hint information for the specified object categories throughout the entire video, serving as a continuous and fixed pseudo-memory source.
However, to ensure optimal results, we still need to carefully adjust the maximum capacity of the memory bank in practical scenarios to determine the ideal value.
This process can also be automated through a script designed to evaluate the best combination of parameters.

To maintain constant GPU memory usage, we have constrained the size of the memory bank. However, the question arises: is it possible to achieve constant memory usage while retaining comprehensive memory information? A potential direction for future research could involve rethinking the design of the memory bank by adopting architectures such as RWKV\cite{rwkv}, which utilize weight-based state representations. This approach may facilitate multi-frame memory storage within a fixed parameter space.

\textbf{3}. One notable issue is the conceptual gap between the detection model and SAM2 during inference.
The detection model outputs categories, and multiple objects of the same category can exist within a single frame.
Conversely, SAM2 receives object IDs, and each object ID corresponds to only one object in a frame.
A limitation of our Det-SAM2 is that it can only be used in scenarios where each category can only appear once in each frame. 
For example, in a pool game scenario, the detection model's output needs to treat each ball as a separate category, ensuring that each ball has a unique object ID when passed to SAM2.
However, if our detection model only distinguishes between the cue ball and other balls, or only between solid and striped balls, SAM2 will face the problem of receiving multiple different prompts for the same object ID at different positions, which confuses SAM2.

We have not yet resolved the gap between the detection model's output categories and SAM2's concept of receiving object IDs in Det-SAM2.
Some potential solutions involve adding engineering checks to manually assign unique and fixed SAM2 object IDs to different objects of the same detection category.
However, this raises the challenge of how to differentiate between various objects of the same detection category to ensure that their IDs remain consistently matched.

\section{Conclusion}
This article presents the implementation process of Det-SAM2, a framework based on SAM2 that extends the system's capabilities without requiring manual interaction.
We have implemented essential features for SAM2 to be applied in specific business scenarios, facilitating automatic SAM2 inference in long videos while maintaining constant GPU memory usage and overall memory efficiency.
We anticipate that additional applications based on SAM2 will emerge in the near future.

\section*{Acknowledgments}
We acknowledge the support provided by Motern AI.

\section*{Author Contributions}
\textbf{Zhiting Wang} : Algorithm team leader. \textbf{Qiangong Zhou} : Core code contribution. \textbf{Zongyang Liu} : Project supervision.

%Bibliography
\bibliographystyle{unsrt}  
\bibliography{template}  

\appendix
\section{Release Old Frames}
\label{A1}
In the \texttt{sam2.sam2\_video\_predictor} module, the \texttt{SAM2VideoPredictor.release\_old\_frames()} method clears the old frames and specifically releases the non-condition frames in \texttt{output\_dict} and \texttt{output\_dict\_per\_obj}, as well as the condition frames in \texttt{output\_dict}, \texttt{output\_dict\_per\_obj}, and \texttt{consolidated\_frame\_inds}.

\begin{verbatim}
def release_old_frames():
    ...
    # delate old non_cond_frames
	inference_state['output_dict']['non_cond_frame_outputs'].pop(old_idx)
	for obj in inference_state['output_dict_per_obj'].keys():
		inference_state['output_dict_per_obj'][obj]['non_cond_frame_outputs'].pop(old_idx)
	# delate old cond_framse
	inference_state['output_dict']['cond_frame_outputs'].pop(old_idx)
	inference_state['consolidated_frame_inds']['cond_frame_outputs'].discard(old_idx)
	for obj in inference_state['output_dict_per_obj'].keys():
		inference_state['output_dict_per_obj'][obj]['cond_frame_outputs'].pop(old_idx)
    ...
\end{verbatim}

\section{Release GPU memory after Memory Attention}
\label{A2}
In the \texttt{sam2.modeling.sam2\_base} module, we add a manual memory release operation after the Memory Attention calculation in the \texttt{SAM2Base.\_prepare\_memory\_conditioned\_features()} method.

\begin{verbatim}
def _prepare_memory_conditioned_features():
	...
    pix_feat_with_mem = self.memory_attention(
        curr=current_vision_feats,
        curr_pos=current_vision_pos_embeds,
        memory=memory,
        memory_pos=memory_pos_embed,
        num_obj_ptr_tokens=num_obj_ptr_tokens,
    )
    # Add release GPU memory
    torch.cuda.empty_cache()
    ...
\end{verbatim}

\section{Post-processing Example}
\label{B1}
As shown in Figure \ref{fig:fig1}, post-processing is a necessary step for Det-SAM2 to move towards higher-level applications.
We have implemented a post-processing example in the billiard scene to demonstrate the potential of our Det-SAM2 in practical applications.
Our post-processing example primarily designs three event detection algorithms for billiard scenes, used to determine: goals, ball-to-ball collisions, and ball rebounds off the table edges.

Specifically, in \texttt{postprocess\_det\_sam2.py}, we first calculate the centroid of each segmentation mask (i.e., the position coordinates of each ball) and the velocity vector of the ball between every two frames based on the masks.
Using the position coordinates and velocity vectors as the foundation, we perform mid-level event detection, such as goals, collisions, and pocket bounces.

\subsection{Goal Detection}
First, obtain the positions of the six pockets from the SAM2 inference backbone's detection model, assign names to the six pockets, and determine which position corresponds to which pocket.

During the traversal of each frame, the following conditions are checked:
\begin{enumerate}
    \item The ball's position in the previous frame is near a pocket, and the ball disappears in the current frame.
    \item The ball's velocity in the previous frame points towards the pocket.
\end{enumerate}
If both conditions are satisfied, it is determined that the ball has entered the target pocket.

\textit{Correction Mechanism}: If the same ball is detected entering a pocket again in subsequent frames, the latest goal information will overwrite the previous record.

\subsection{Ball Collision Detection}
During the traversal of each frame, collision detection is triggered when a ball's velocity vector undergoes a significant change (exceeding a defined threshold). 
The following conditions are checked:
\begin{enumerate}
    \item Identify the ball that might have collided with the current ball by analyzing the velocity vectors before and after the event:
    \begin{enumerate}
        \item Before the collision, the two balls are moving towards each other.
        \item After the collision, the velocities of the two balls change significantly, and their accelerations exhibit correlations (e.g., the introduction of components indicating they are moving away from each other).
    \end{enumerate}
    \item Determine if the potential collision ball is near the current ball.
\end{enumerate}
If both conditions are satisfied, it is determined that a collision occurred between the two balls.

\textit{Correction Mechanism}: If the same frame is reevaluated later and yields a different result, the new judgment will overwrite the previous information.

\textit{Note}: Collision detection requires acceleration calculations, which necessitate data from the current frame and the two preceding frames, totaling three frames of information.

\subsection{Table Edge Rebound Detection}
First, extract the four valid boundaries of the table (top, bottom, left, right) from the coordinates of the six pockets.
Shrink these boundaries inward to create buffer zones near each edge, which are used to trigger rebound detection.

During the traversal of each frame, when a ball enters a buffer zone near the boundaries, the boundary position (top, bottom, left, or right) is recorded, and rebound detection is triggered based on the following conditions:
\begin{enumerate}
    \item Check if the ball was moving toward the corresponding boundary before the rebound (in the previous frame).
    \item Check if the ball is moving away from the corresponding boundary after the rebound (in the current frame).
    \item Verify whether the velocity component perpendicular to the boundary has essentially reversed direction. If not, check if the velocity component parallel to the boundary remains approximately consistent.
\end{enumerate}
If conditions 1, 2, and 3 are all satisfied, it is determined that the ball rebounded off the corresponding boundary.
If condition 1 is satisfied but conditions 2 and 3 are not, further checks are performed to determine if the irregular behavior is due to the ball hitting a curved surface near a pocket:
\begin{enumerate}
    \item Check if the ball is near a pocket.
    \item Determine if the velocity vector has changed significantly between the current frame and the previous frame (indicating a possible external collision).
    \item Confirm that the velocity vector in the previous frame is not directed toward any other ball (to rule out ball-to-ball collisions near the pocket).
    \item Verify that there are no collisions involving the ball in the current frame (using the collision results dictionary).
\end{enumerate}
If conditions 4, 5, 6, and 7 are all satisfied, it is also determined that the ball rebounded off the corresponding boundary.

\textit{Correction Mechanism}: If the same frame is reevaluated later with inconsistent results, the new judgment will overwrite the previous one.

\textit{Note}: Rebound detection requires velocity calculations, which depend on data from the current frame and the two preceding frames, totaling three frames of information.

\section{Det-SAM2-pipeline}
\label{B2}
The Det-SAM2-pipeline is a complete workflow that integrates Det-SAM2 with post-processing.

The script \texttt{Det-SAM2\_pipeline.py} utilizes the video inference backbone class from \texttt{det\_sam2\_RT.py} and the post-processing class from \texttt{postprocess\_det\_sam2.py}, combining them within the \texttt{DetSAM2Pipeline.inference()} function.

During the execution of \texttt{DetSAM2Pipeline.inference()}, the SAM2 video inference backbone and the post-processing components each operate in separate threads, enabling an asynchronous and parallel workflow:
\begin{enumerate}
    \item Main Inference Thread:
    Responsible for reading data frame by frame from the video stream and performing detection and segmentation inference.
    \begin{enumerate}
        \item Reads each frame from the video stream as input from the video source.
        \item Passes the frames into the Det-SAM2 inference framework, where the detection model provides conditional prompts for SAM2 to perform segmentation and segmentation correction in a real-time video stream.
        \item Stores the segmentation results of each inference (\texttt{propagate\_in\_video}) into the inference result cache \texttt{video\_segments}. Newly added inference results are also pushed to the post-processing queue (\texttt{frames\_queue}).
        \item Triggers the post-processing thread once the required settings (e.g., pocket coordinates, table boundaries) are collected, ensuring the post-processing thread is activated.
    \end{enumerate}
    \item Post-Processing Thread:
    Handles the segmentation results pushed by the main inference thread in parallel, performing further object tracking and state analysis.
    \begin{enumerate}
        \item Monitors the post-processing queue (\texttt{frames\_queue}), and starts processing as soon as new inference results are available. It processes all frames sequentially (may reprocess previously processed frames) but does not skip frames to directly process later ones.
        \item Uses the segmentation results retrieved from \texttt{frames\_queue} to calculate the ball positions and velocity vectors.
        \item Performs goal detection starting from frame 2.
        \item Performs collision detection starting from frame 3.
        \item Performs rebound detection starting from frame 3.
    \end{enumerate}
\end{enumerate}
This two-thread asynchronous parallel workflow ensures the Det-SAM2 system is efficient, accurate, and capable of real-time processing.

\end{document}